\documentclass{article}

\usepackage{PRIMEarxiv}

\usepackage[utf8]{inputenc} 
\usepackage[T1]{fontenc}    
\usepackage{hyperref}       
\usepackage{url}            
\usepackage{booktabs}       
\usepackage{amsfonts}       
\usepackage{nicefrac}       
\usepackage{microtype}      
\usepackage{lipsum}
\usepackage{fancyhdr}       
\usepackage{graphicx}       
\usepackage{hyperref}       
\usepackage{appendix}
\usepackage{bm}             
\usepackage{float}
\usepackage[symbol]{footmisc} 
\usepackage{fontenc}
\graphicspath{{media/}}     

\title{\textbf{\textbf{ADRD}}: LLM-Driven \textbf{A}utonomous \textbf{D}riving Based on \textbf{R}ule-based \textbf{D}ecision Systems
}

\author{
  \begin{tabular}{cc}
    Fanzhi Zeng\textsuperscript{\footnotemark[1]} & 
    Siqi Wang\textsuperscript{\footnotemark[1]} \\
    \texttt{zengfz21@mails.tsinghua.edu.cn} & 
    \texttt{wang-sq24@mails.tsinghua.edu.cn} \\
    \\
    Chuzhao Zhu& 
    Li Li\textsuperscript{\footnotemark[2]} \\
    \texttt{zhucz21@mails.tsinghua.edu.cn} & 
    \texttt{li-li@mail.tsinghua.edu.cn}
  \end{tabular}
}

\begin{document}
\maketitle

\footnotetext[1]{Co-first author}
\footnotetext[2]{Corresponding author}

\begin{abstract}



How to construct an interpretable autonomous driving decision-making system has become a focal point in academic research. In this study, we propose a novel approach that leverages large language models (LLMs) to generate executable, rule-based decision systems to address this challenge. Specifically, harnessing the strong reasoning and programming capabilities of LLMs, we introduce the \textbf{ADRD}(LLM-Driven \textbf{A}utonomous \textbf{D}riving Based on \textbf{R}ule-based \textbf{D}ecision Systems) framework, which integrates three core modules: the Information Module, the Agents Module, and the Testing Module. The framework operates by first aggregating contextual driving scenario information through the Information Module, then utilizing the Agents Module to generate rule-based driving tactics. These tactics are iteratively refined through continuous interaction with the Testing Module. Extensive experimental evaluations demonstrate that \textbf{ADRD} exhibits superior performance in autonomous driving decision tasks. Compared to traditional reinforcement learning approaches and the most advanced LLM-based methods, \textbf{ADRD} shows significant advantages in terms of interpretability, response speed, and driving performance. These results highlight the framework’s ability to achieve comprehensive and accurate understanding of complex driving scenarios, and underscore the promising future of transparent, rule-based decision systems that are easily modifiable and broadly applicable. To the best of our knowledge, this is the first work that integrates large language models with rule-based systems for autonomous driving decision-making, and our findings validate its potential for real-world deployment.

\end{abstract}

\section{Introduction}

The rapid development of autonomous driving technology has placed increasingly higher demands on the interpretability of decision-making systems. Achieving a "white-box" autonomous driving decision model, where the internal logic of the decision-making process is transparent and comprehensible to humans, has become a focal research direction in academia\cite{chen2021interpretable, wang2023chatgpt, wen2023dilu, chen2024driving, xu2024drivegpt4}. In recent years, with the advancement of deep learning, neural network-based decision systems have emerged as the dominant approach in autonomous driving due to their data-driven nature and ease of learning. However, such systems typically rely heavily on training data, leading to a sharp decline in performance when encountering out-of-distribution driving scenarios. Moreover, these models are often over-parameterized, making their decision logic difficult to interpret and challenging to modify or debug by domain experts when suboptimal performance occurs.


Rule-based decision systems, on the other hand, hold significant value in the field of interpretable autonomous driving due to their intrinsic transparency and modifiability. By defining explicit rules to guide vehicle behavior, these models offer strong traceability and explainability. Nevertheless, traditional rule-based systems are highly dependent on expert knowledge, resulting in high development costs and limited adaptability to complex and dynamic traffic environments. Furthermore, their structural characteristics make it difficult to integrate data-driven learning methods, thereby hindering automated improvement as new scenarios emerge, and making the construction of high-complexity rule-based models prohibitively expensive.


Recently, the emergence of large language models (LLMs)\cite{radford2018improving, radford2019language, mann2020language, achiam2023gpt, guo2025deepseek} opens up new possibilities for rule-based autonomous driving decision-making. Trained on vast corpora, LLMs exhibit extensive world knowledge and strong reasoning capabilities, enabling them to automatically generate logically coherent rule sets without human intervention. This capability offers a novel pathway toward building highly interpretable, rule-based decision models. Although some LLM-based approaches for interpretable autonomous driving decisions have been proposed\cite{wang2023chatgpt, wen2023dilu, chen2024driving, xu2024drivegpt4}, they remain largely model-centric, which introduces several key limitations:


\begin{quote}
\textbf{Low response efficiency:} Most existing methods operate by generating a single decision based on a single-frame input, resulting in slow overall response times that fail to meet the strict real-time requirements of autonomous driving systems. Additionally, this local perspective limits comprehensive scene understanding and unified modeling.

\textbf{Poor tactic flexibility:} Even though some methods can produce interpretable decision-making processes, their underlying mechanism remains data-driven fine-tuning, which lacks high-level abstraction of driving scenarios. Consequently, such tactics struggle to generalize effectively to novel or extreme driving situations. Moreover, despite the availability of interpretable outputs, researchers still find it difficult to directly modify the driving policy model to meet diverse requirements.
\end{quote}


To address these challenges, we propose a novel framework \textbf{ADRD}(LLM-Driven \textbf{A}utonomous \textbf{D}riving Based on \textbf{R}ule-based \textbf{D}ecision Systems) that leverages large language models to automatically generate rule-based driving policies, aiming to achieve efficient, interpretable, and robust decision-making in autonomous driving. Decision trees, as a classical rule-based model, naturally offer both interpretability and fast inference speed, making them well-suited for various driving scenarios. Their structured representation also facilitates policy transfer and modification. Therefore, in this work, we adopt decision trees as the concrete implementation of our rule-based autonomous driving system. Specifically, our framework consists of three core modules: the Planner, the Coder, and the Summarizer, forming a closed-loop agent system. First, the Planner generates high-level strategic descriptions based on the input driving scenario description, predefined rules, and vehicle state. Then, the Coder translates these descriptions into executable decision tree code. Finally, after simulation validation, failure cases are fed into the Summarizer module for iterative optimization of the tactic, which is then fed back to the Planner. Notably, we use executable code as the medium for interaction between the decision tree and the driving environment, a design choice whose advantages have been demonstrated in \cite{wang2024executablecodeactionselicitcodeact, deng2025smacr1emergenceintelligencedecisionmaking, lin2024rlllmdtautomaticdecisiontree}.

\begin{figure}[htbp]
    \centering
    \includegraphics[width=1.0\textwidth]{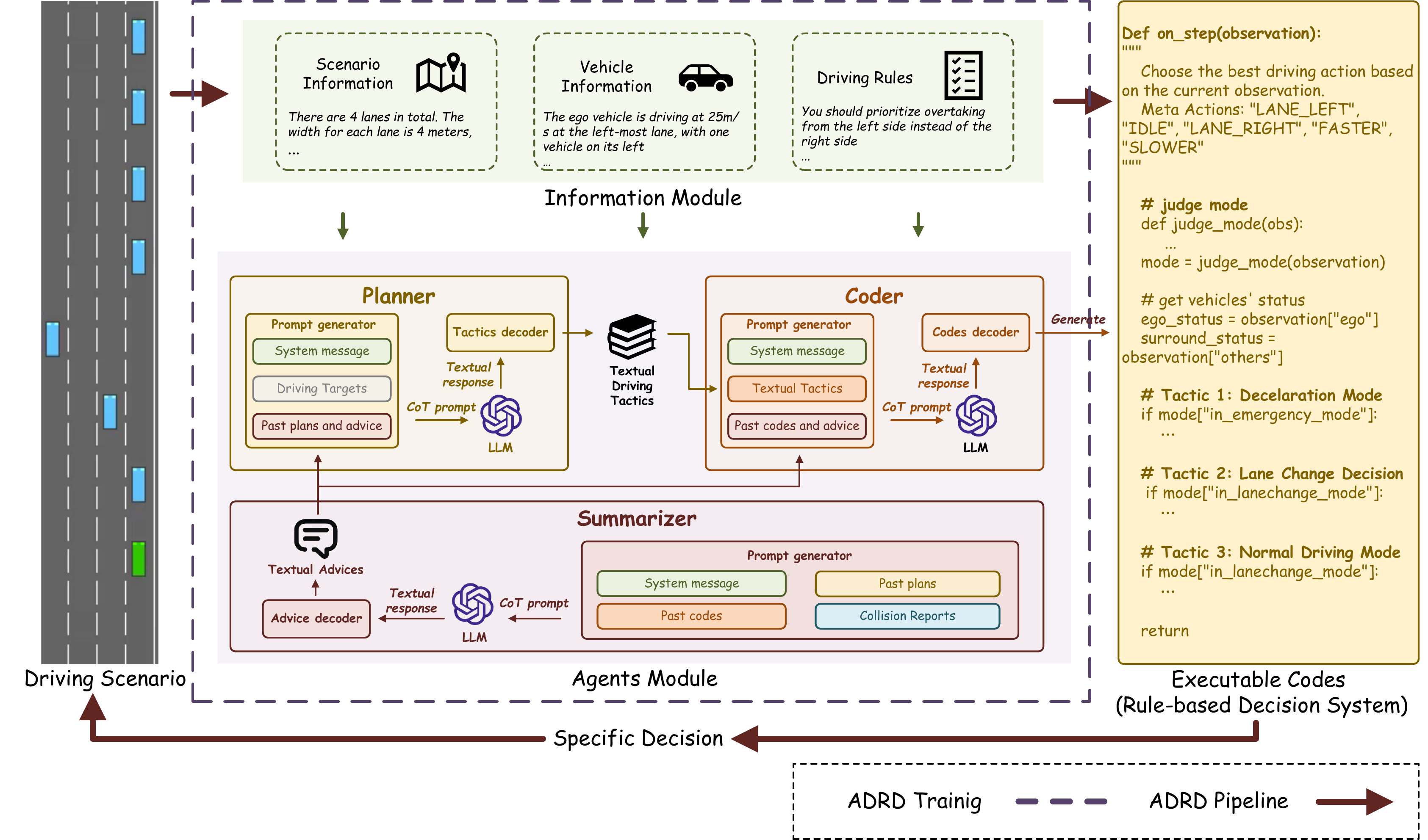}
    \caption{Overall Stucture of \textbf{ADRD}.}
    \label{fig:ADRD_OVERALL}
\end{figure}


We evaluate the performance of \textbf{ADRD} on \textbf{highway-env}, a popular autonomous driving simulation platform. Extensive experimental results demonstrate that \textbf{ADRD} exhibits strong generalization and robustness across multiple typical driving scenarios. Compared to traditional knowledge-driven approaches and modern data-driven reinforcement learning methods, \textbf{ADRD} achieves significant improvements in decision-making performance, response efficiency, and interpretability, highlighting its great potential for deployment in real-world autonomous driving systems.




This paper is organized as follows: \textbf{Section 2} reviews related work; \textbf{Section 3} presents the detailed framework design of \textbf{ADRD}; \textbf{Section 4} reports experimental results under different driving scenarios and parameter settings, along with comparisons against baseline methods; and \textbf{Section 5} concludes the paper. Our code is publicly available at \url{https://github.com/bjbcjwsq/ADRD}.

\section{Related Works}




\textbf{Rule-Based Decision Systems:} Rule-based decision systems make decisions using explicitly defined rules, which endows them with strong interpretability and modifiability. Classic rule-based decision algorithms include rule engines\cite{davis1975production}, expert systems\cite{liao2005expert}, and fuzzy logic\cite{zadeh1965fuzzy}. Among these, decision trees\cite{magee1964decision, rokach2005decision, kotsiantis2013decision} represent a canonical approach to rule-based decision modeling. They recursively partition a dataset into smaller subsets to construct a tree-like structure used for decision-making or prediction. Each internal node represents a test on a specific feature, while each leaf node corresponds to a final class label or output value. This hierarchical structure offers high transparency and interpretability. Traditionally, the construction of decision trees has been approached through two main paradigms: knowledge-driven methods—primarily relying on human experts—and data-driven approaches, such as classical machine learning algorithms like CART\cite{loh2011classificationCART} and ID3\cite{quinlan1986inductionID3}. However, these conventional methods suffer from several limitations, including heavy dependence on expert experience, susceptibility to expert bias, and sensitivity to training data, all of which pose significant challenges when applying decision trees to modern complex systems. Recent advances in large language models (LLMs), however, have opened up new possibilities for knowledge-driven decision tree construction. For example, \cite{deng2025smacr1emergenceintelligencedecisionmaking} have used large models to generate decision trees for zero-sum games in StarCraft, preliminarily validating the feasibility of this paradigm. Further work \cite{lin2024rlllmdtautomaticdecisiontree} combined LLM-generated decision trees with reinforcement learning, enabling co-evolution between RL agents and LLMs through mutual interaction. These emerging studies highlight the promising potential of using LLM-generated decision trees for interpretable decision-making.


\textbf{Code Generation for Decision-Making Using Large Language Models:} In tasks involving the construction of decision trees using large models, the generated decision trees are typically returned as Python code blocks\cite{deng2025smacr1emergenceintelligencedecisionmaking, lin2024rlllmdtautomaticdecisiontree}. This implies a significant demand on the coding and reasoning capabilities of large language models. Recently, large language models have made remarkable progress in code generation. For instance, Codex\cite{chen2021evaluatingcodex} based on GPT-3\cite{mann2020language} and PyCodeGPT\cite{CERTpycodegpt} based on GPT-Neo\cite{gpt-neo}, have been trained on large-scale, high-quality code datasets crawled from the web, enabling them to outperform general-purpose language models on specialized code generation tasks. Additionally, several recent studies have explored the advantages of using executable code as a unified action space for LLM agents. For instance, the CodeAct\cite{wang2024executablecodeactionselicitcodeact} framework proposes using executable Python code as the primary form of decision-making and demonstrates its superiority over traditional formats such as JSON or plain text. This successful case suggests that having large models generate executable code as part of the action space offers a promising direction for decision-making in complex systems like autonomous vehicles.


\textbf{Interpretable Autonomous Driving Decisions}: Interpretable autonomous driving decisions aim to break open the black box of decision models so that humans can understand the underlying logic of decisions. This is crucial for improving transparency, trustworthiness, and the ethical and legal analysis of autonomous systems. Before the advent of large language models, interpretable autonomous driving decisions were typically achieved through methods such as decision tree construction or imitation learning\cite{atakishiyev2024explainable}. While these approaches allowed models to provide rudimentary explanations for their decisions, their explanatory power was limited by weak semantic understanding. With the emergence of large language models, detailed comprehension of driving decisions has become possible. For example, Dilu\cite{wen2023dilu} designed a closed-loop decision agent composed of memory, reasoning, and reflection modules by modeling the thinking process of human drivers, achieving efficient scene understanding and driving decisions without model fine-tuning. Other works, such as DriveGPT4\cite{xu2024drivegpt4}, adopted multimodal fine-tuning to allow language models to simultaneously generate both driving decisions and corresponding textual explanations. However, these works still focus primarily on individual scenarios without considering the overarching logical principles of driving. This narrow perspective limits both their response efficiency and their ability to understand macro-level features across similar driving scenarios.

\section{Methodology}


In this section, we present a comprehensive overview of the \textbf{ADRD} framework and detailed implementation of each submodule within the Agents Module. The overall architecture of \textbf{ADRD}, as shown in Figure~\ref{fig:ADRD_OVERALL}, consists of three primary modules: \textbf{Information Module}, \textbf{Agents Module}, and \textbf{Testing Module}. These modules are responsible for constructing an initial driving information set, analyzing and reasoning about driving decision logic, and conducting testing and feedback, respectively. Among them, the Agents Module is the core of \textbf{ADRD} and is further divided into \textbf{Planner}, \textbf{Coder}, and \textbf{Summarizer}, which are responsible for generating preliminary driving tactics, converting these tactics into executable Python code, and improving driving tactics based on simulation failure results, respectively. Through a well-designed prompt engineering framework, \textbf{ADRD} can continuously refine its decision tree while maintaining safe driving and good interpretability. In the following, we employ decision trees as the concrete implementation of the rule-based decision system.

\subsection{Overview}

\begin{figure}[htbp]
    \centering
    \includegraphics[width=1.0\textwidth]{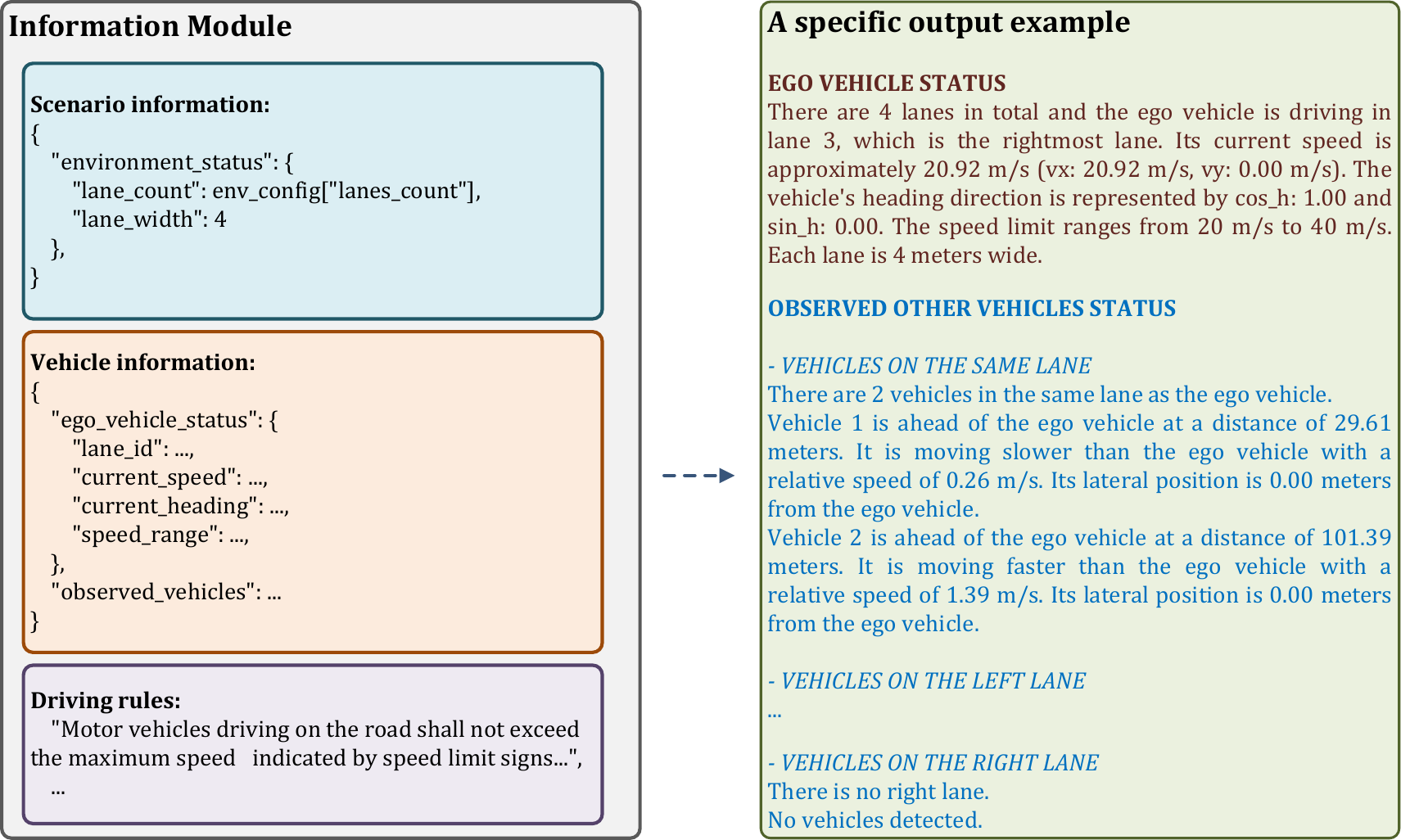}
    \caption{specific Output Example of the \textbf{Information Module}.}
    \label{fig:INFORMATION_MODULE}
\end{figure}

We first introduce the training process and pipeline of \textbf{ADRD}. For the training process of \textbf{ADRD}, at the beginning, the \textbf{Information Module} converts scenario information, vehicle information, and driving rules into natural language text understandable by LLMs. Driving environment and vehicle information comes from the vehicle's environmental observations; in this paper, we assume it to be highly abstracted environmental features such as the number and width of lanes, the position and speed of the ego-vehicle, etc. Driving rules come from predefined traffic regulations, establishing a basic understanding of driving ethics for LLMs. A specific output example of the \textbf{Information Module} is shown in Figure~\ref{fig:INFORMATION_MODULE}, enabling the LLM to initially grasp the macroscopic understanding of the driving scenario, laying the foundation for comprehension, reasoning, and decision tree construction in the \textbf{Agents Module}.

The refined information from the \textbf{Information Module} is then fed into the \textbf{Agents Module}. This module is the core component of \textbf{ADRD}, where the \textbf{Planner}, \textbf{Coder}, and \textbf{Summarizer} collaborate effectively. They not only ensure robust forward reasoning in the decision tree but also improve it based on failed interactions, achieving effective closed-loop iteration. Specifically, first, the \textbf{Planner} generates preliminary driving tactics based on textual descriptions of the driving scenario and predefined driving targets (e.g., conservative and safe driving or efficient and fast driving). Second, the \textbf{Coder} translates the detailed driving tactic description from the \textbf{Planner} into executable Python code suitable for a specific simulation environment. This code is deployed into real or simulated driving environments for validation. Finally, the \textbf{Summarizer} evaluates the tactics generated by the \textbf{Planner} and the code produced by the \textbf{Coder} based on collision reports from the testing module, identifies potential issues, and provides detailed improvement suggestions for both modules. By iterating through this closed-loop control process repeatedly, the \textbf{Agents Module} continually enhances its understanding and control capabilities in driving scenarios, resulting in increasingly better-performing decision trees.


For the pipeline of \textbf{ADRD}, \textbf{ADRD} takes the driving information extracted by the \textbf{Information Module} from the driving environment and feeds it into the well-trained rule-based decision system to generate specific driving decisions. These decisions are then executed in the driving environment to interact with the current scene and obtain the next frame of the driving scenario. This process continues in a loop until the preset maximum driving time is reached or a hazardous driving event occurs.

Next, we will detail how each of the three expert sub-modules in the \textbf{Agents Module} operates.

\subsection{\textbf{Planner}}


During the generation of the driving behavior decision tree, the \textbf{Planner} primarily assumes the roles of cognition, reasoning, and planning. Compared to directly generating executable decision tree Python code from environmental information, the \textbf{Planner} alleviates some cognitive and reasoning pressure by first generating human-readable textual driving tactics before producing executable code. This not only improves the explainability of the decision tree formation process but also contributes to enhancing its performance.


The prompt generator for the \textbf{Planner} consists of three components: system prompts, driving targets, and past plans along with advice for improvement. The system prompts describe the \textbf{Planner}’s role within the agent group and specify a standardized output format. The driving targets define the desired driving style, such as conservative or aggressive, for which the \textbf{Planner} should construct the decision tree. Historical planning records and improvement suggestions are optional inputs that, when provided, serve as references to help the \textbf{Planner} generate more refined and robust driving tactics. By synthesizing all this information, the \textbf{Planner} analyzes the current driving scenario and proposes a set of driving tactics most relevant to the situation. For each tactic, the \textbf{Planner} defines its name, usage conditions, priority, and detailed execution, facilitating the \textbf{Coder}’s understanding of how to integrate these tactics into a coherent and unified decision tree. Take the tactic named \textbf{"Active Lane Change Operation"} as an example. The execution process of this tactic is:




\begin{quote}

\textbf{1.} Evaluate the distance and closing speed of the preceding vehicle. When the distance is significantly less than the safe distance and the rate of closure is rapid (usage condition):
\begin{quote}
   \textbf{1.a} First, check whether there is sufficient space in the adjacent lane to complete the lane change operation. If the ego-vehicle is not in the leftmost lane, prioritize changing to the left lane; otherwise, consider the right lane.
   
   \textbf{1.b} Second, assess the safety of the lane change to the adjacent lane, considering vehicle speeds and other potential hazards.
   
   \textbf{1.c} If the target lane meets safety conditions, execute the lane change immediately.
\end{quote}
\textbf{2.} If the lane change conditions are not met, fall back to deceleration operations (refer to the tactic "Conservative Deceleration") (priority).

\end{quote}



\begin{figure}[htbp]
    \centering
    \includegraphics[width=1.0\textwidth]{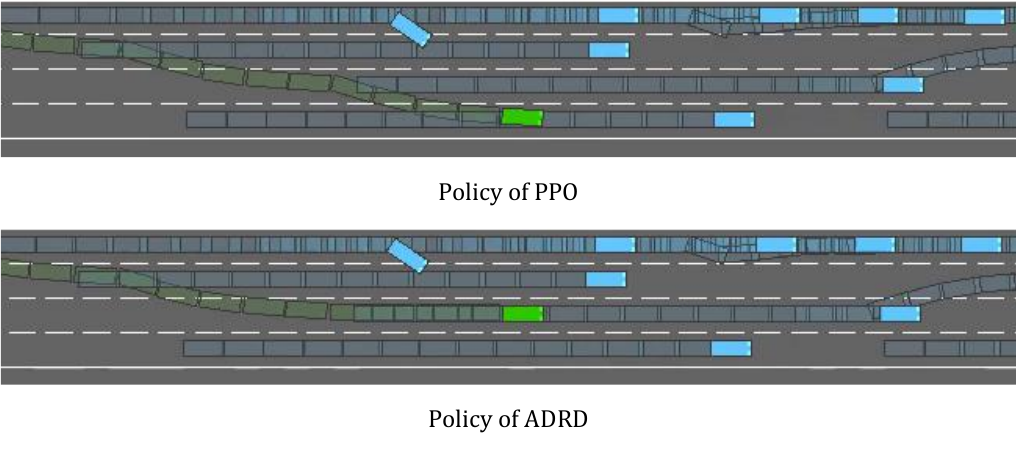}
    \caption{Policy of PPO vs Policy of \textbf{ADRD}.}
    \label{fig:RLVSADRD}
\end{figure}

Generating human-readable, interpretable tactics offers several key advantages. First, it enables human experts to understand the intermediate process by which large language models (LLMs) generate decision trees. More importantly, it allows for direct modification of the tactic content, facilitating alignment with real-world driving requirements. Second, this approach can fully leverage the pre-trained knowledge and common-sense reasoning capabilities of LLMs, ensuring that the generated driving policies are grounded in human-like understanding and behavior. Figure~\ref{fig:RLVSADRD} illustrates a comparison between the driving tactics generated by a reinforcement learning (RL) algorithm and those produced by an LLM in a particular high-speed driving scenario. Although both tactics achieve relatively high safety rewards, the RL-generated policy chooses to switch to the more hazardous fourth lane, abandoning the relatively safer third lane, which is inconsistent with normal human driving behavior. Moreover, due to the black-box nature of neural network-based policies, it is difficult for human experts to directly adjust the learned parameters to correct such undesirable behaviors. In contrast, the LLM-generated tactic not only aligns naturally with human driving conventions—choosing to follow traffic more safely in the appropriate lane, but also remains easily interpretable and modifiable by human experts. These results demonstrate that \textbf{ADRD} effectively overcomes the limitations of traditional decision-making methods, particularly their lack of transparency and difficulty in manual correction. Detailed examples of prompts are available in Appendix A.1.

\subsection{\textbf{Coder}}


The main responsibility of the \textbf{Coder} is to generate executable Python functions for the simulation environment based on textual driving tactics generated by the \textbf{Planner}. The \textbf{Coder} relies on the strong code-writing capabilities of current LLMs, as numerous studies have demonstrated the promising applications of LLM-generated code as an action space. The \textbf{Coder}’s prompt generator also comprises three components: system prompts, textual driving tactic descriptions, and past code along with improvement suggestions. System prompts define the \textbf{Coder}’s role and standardized output format within the agent group, including detailed descriptions of variable meanings and formats in the observation space to ensure the generated code can be decoded programmatically. The textual driving tactic descriptions come from the \textbf{Planner}’s output, while past code and improvement suggestions originate from the \textbf{Summarizer}, similar to the \textbf{Planner}. Based on this information, the \textbf{Coder} generates executable Python code—in particular, as a Python function taking observations as input, which is fed into the simulation environment for interaction. This output format facilitates the integration of textual driving tactics into the simulation environment for controlling the ego-vehicle, greatly simplifying the self-iteration process of \textbf{ADRD}. Detailed prompt examples are provided in Appendix A.2.


Figure~\ref{fig:BRIEF_TREE} further visualizes a converged driving tactic function as a decision tree structure, making it easy for human experts to read, analyze, and identify potential logical vulnerabilities. This significantly enhances the reliability and transparency of intermediate steps in autonomous driving decisions. Notably, by incorporating different constraints into the prompts of the \textbf{Planner} and \textbf{Coder}, such as "driving style descriptions", \textbf{ADRD} can generate decision trees of varying complexity, a topic discussed in detail in the "Experiments" section.

\subsection{Summerizer}


Although the \textbf{Planner} and \textbf{Coder} have completed the initial construction of the autonomous driving decision tree, to ensure the decision tree is reliably usable in driving environments, appropriate regulatory and feedback mechanisms must be introduced to safeguard driving efficiency and safety—this is the primary function of the \textbf{Summarizer}. The \textbf{Summarizer}’s prompt generator includes three components: system prompts, collision reports, and corresponding decision tree code implementations. System prompts describe the responsibilities of the \textbf{Summarizer}, namely identifying whether errors reside in the driving tactic itself generated by the \textbf{Planner} or in the Python code implementation by the \textbf{Coder}. Collision reports are formatted by the \textbf{Testing Module} and include historical trajectories of the ego-vehicle and surrounding vehicles over the past T time steps, serving as reference data for the LLM to infer collision causes and map them to problems in driving tactics and executable codes. The driving tactics and corresponding decision tree codes were generated by the \textbf{Planner} and \textbf{Coder} in the previous iteration step, assisting the LLM in locating issues and generating recommendations for improvement in the respective modules. 


Designing the \textbf{Summarizer} as an independent unit rather than integrating it into the \textbf{Planner} and \textbf{Coder} modules offers several advantages: first, it avoids requiring both modules to reflect after every collision, saving training time costs and computational resources for LLM inference; second, it promotes specialization among different modules, effectively reducing the complexity faced by the LLM in handling individual tasks, lowering inference load, and improving LLM performance on both single and overall tasks. Among the three sub-modules in the \textbf{Agents Module}, the \textbf{Summarizer} undertakes the majority of cognitive and reasoning tasks, serving as the core component enabling \textbf{ADRD}’s self-iterative evolution. Detailed prompt examples are available in Appendix A.3.

\begin{figure}[htbp]
    \centering
    \includegraphics[width=1.0\textwidth]{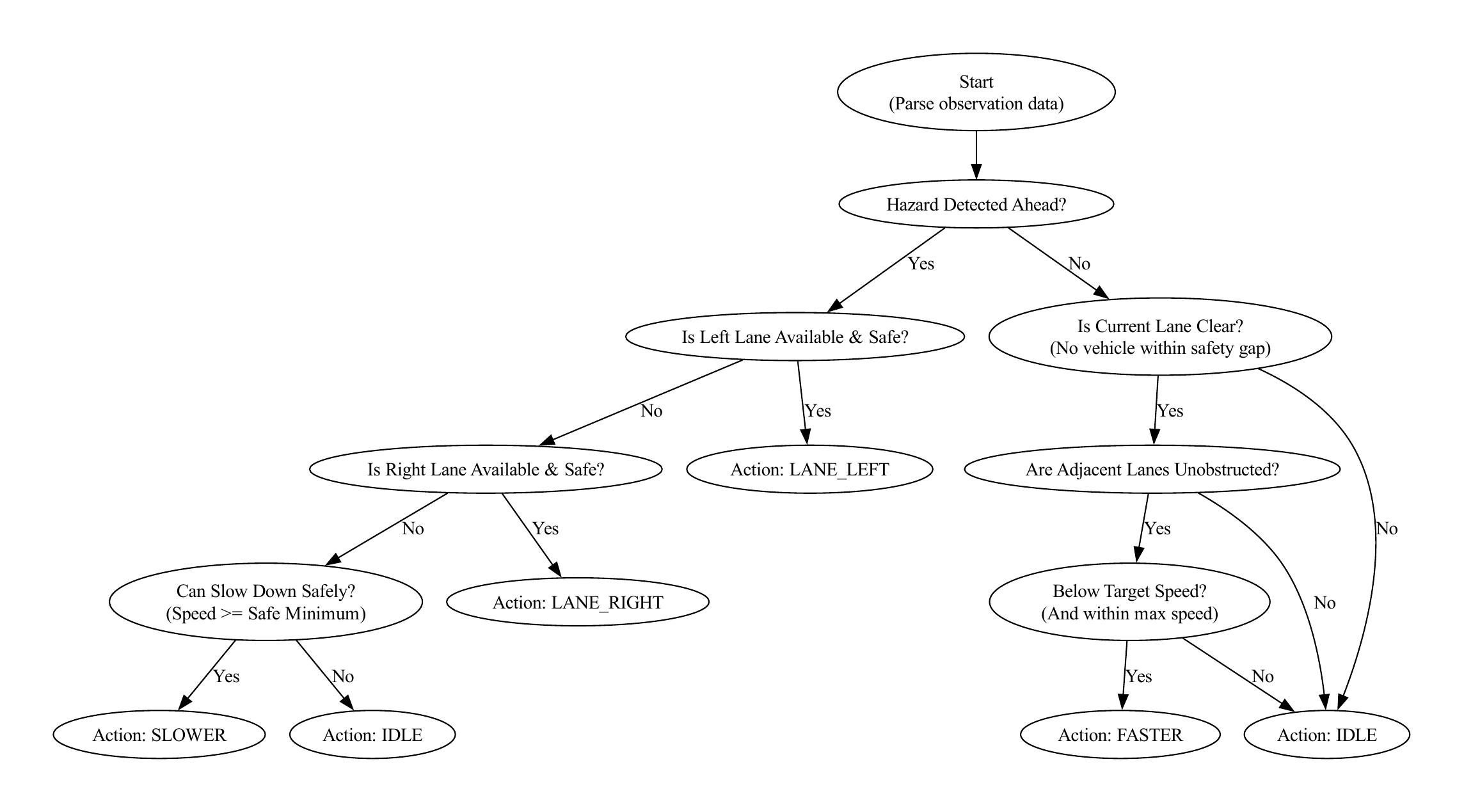}
    \caption{A converged decision tree generated by \textbf{ADRD}. To concisely illustrate the key structure of the decision tree, some specific numerical values have been omitted. For a more detailed version of the decision trees, please refer to the Experiment section.}
    \label{fig:BRIEF_TREE}
\end{figure}

\section{Experiments}

\subsection{Settings and Baselines}


We conduct experiments using the \textbf{highway-v0} scenario from the \textbf{highway-env} environment. This scenario simulates an autonomous driving environment on a highway, which we use to compare the performance of \textbf{ADRD} with several baseline autonomous driving decision-making methods. The baselines include PPO-CLIP\cite{schulman2017proximal}, a representative algorithm in the field of reinforcement learning (RL), and Dilu\cite{wen2023dilu}, a knowledge-driven decision-making method based on large language models. Among all the methods evaluated on highway-env, DiLu demonstrates the best decision-making performance, to the best of our knowledge.


In all autonomous driving policies, we adopt the discrete action space provided by \textbf{highway-v0} as the set of possible driving decisions for the ego vehicle. These actions include:
\begin{itemize}
    \item Maintain current speed and lane (\textbf{IDLE}),
    \item Decelerate (\textbf{SLOWER}),
    \item Accelerate (\textbf{FASTER}),
    \item Change lane to the left (\textbf{LANE\_LEFT}),
    \item Change lane to the right (\textbf{LANE\_RIGHT}).
\end{itemize}

The decision-making frequency is set to 1Hz. For speed control, there are five speed levels: 20m/s, 25m/s, 30m/s, 35m/s, and 40m/s. Executing acceleration or deceleration actions allows the ego vehicle to shift between these speed levels. If the vehicle is already at the lowest or highest speed level, further deceleration or acceleration has no effect on its speed.\\

\textbf{PPO-CLIP}


We employ PPO-CLIP as a representative RL baseline. PPO-CLIP is a widely used policy gradient algorithm in deep reinforcement learning. Its core idea is to introduce a "trust region" during policy updates, limiting the divergence between the new and old policies to prevent training instability. This method offers advantages such as parameter insensitivity, ease of tuning, and high sample efficiency. The exact form of its objective function is detailed in the original paper~\cite{schulman2017proximal}, and is omitted here for brevity.


In our experimental setup, the observation space includes state information from the ego vehicle and up to 7 of the nearest surrounding vehicles, including their positions, velocities, and headings. The policy network consists of a two-layer multi-layer perceptron (MLP), with each layer containing 128 neurons. The learning rate is set to $3 \times 10^{-4}$, and the model is trained for a total of 90,000 action steps. This configuration enables PPO-CLIP to serve effectively as a representative of traditional RL-based autonomous driving decision methods, allowing us to compare its performance with that of ADRD and Dilu.\\


Both Dilu and ADRD are trained using OpenAI's \textbf{o3-mini}\cite{el2025o3} model, which is specifically designed to offer fast and cost-effective reasoning capabilities, excelling particularly in tasks related to programming, mathematics, and science.


\textbf{Dilu}


We select Dilu as a representative autonomous driving decision method leveraging large language models. Dilu simulates the learning process of human drivers through three core modules:
\begin{itemize}
    \item \textbf{Memory Module}: Stores historical driving experiences,
    \item \textbf{Reasoning Module}: Combines the current driving context with a subset of relevant past experiences to generate decisions,
    \item \textbf{Reflection Module}: Evaluates the safety of decisions and updates the memory bank to improve future behavior.
\end{itemize}

In our experiments, we configure Dilu to use the 3 most relevant memory entries for each decision-making step. That is, at every time step, Dilu refers to the 3 historical experiences most similar to the current driving situation to assist in reasoning and decision-making. 

\subsection{Main Results}


For the three methods mentioned in Section 4.1, we evaluate their performance across three driving scenarios with varying levels of difficulty: specifically, a 4-lane environment with a vehicle density of 2.00, a 5-lane environment with a density of 2.50, and a 6-lane environment with a density of 3.00, representing normal, hard, and extreme driving conditions, respectively. The density parameter is provided internally by highway-env and reflects the number of vehicles per unit length of lane.

For each method, we measure two key metrics: the average safe driving time(seconds) over 20 different randomized scenarios, and the average inference time per decision (seconds per command). These metrics capture the driving performance and decision-making efficiency of different autonomous driving approaches, respectively.

It should be noted that both PPO and \textbf{ADRD} perform inference on two Intel(R) Xeon(R) Platinum 8374B CPUs. For PPO, the inference time corresponds to the forward pass of its policy network, while for \textbf{ADRD}, it depends on the complexity of the executable Python functions. DiLu, on the other hand, requires calling OpenAI's \textbf{o3-mini} model for each decision, and its inference time mainly comes from the response speed of the LLM.

The results of the three methods are presented in Table~\ref{tab:MAIN_RESULTS}. It can be seen that, across all three driving scenarios, \textbf{ADRD} not only achieves the longest average safe driving times among all methods but also demonstrates the fastest inference speed. Notably, PPO performs the worst in terms of driving performance despite being trained for 90,000 driving frames, which is much more than the other two methods, highlighting the limitations of reinforcement learning-based data-driven approaches in autonomous driving decision-making.

\begin{table}[h]
\centering
\caption{Comparison of Average Driving Time and Control Efficiency under Different Scenario Settings}
\label{tab:MAIN_RESULTS}
\begin{tabular}{@{}c c c c c@{}}
\toprule
\textbf{Density} & \textbf{Lane Number} & \textbf{Method} & \textbf{Average Driving Time (s)} $\uparrow$ & \textbf{Control Efficiency (s/command)} $\downarrow$ \\
\midrule
& & PPO \cite{schulman2017proximal} & 10.90 & $2.0\times 10^{-4}$ \\
2.00(\textbf{Normal}) & 4 & DiLu \cite{wen2023dilu}      & 23.00                  & 14.33 \\
         & & \textbf{ADRD}                      & \textbf{25.15}  & <\bm{$1.0\times 10^{-6}$ } \\
\midrule
& & PPO \cite{schulman2017proximal} & 8.60 & $2.0\times 10^{-4}$ \\
2.50(\textbf{Hard}) & 5  & DiLu \cite{wen2023dilu}      & 16.00                   & 12.42 \\
            &  & \textbf{ADRD}                      & \textbf{16.75}     & <\bm{$1.0\times 10^{-6}$ } \\
\midrule
 &   & PPO \cite{schulman2017proximal} & 5.50 & $2.0\times 10^{-4}$ \\
3.00(\textbf{Extreme}) & 6 & DiLu \cite{wen2023dilu}      & 10.10                  & 12.67 \\
           & & \textbf{ADRD}                      & \textbf{13.55}    & <\bm{$1.0\times 10^{-6}$ } \\
\bottomrule
\end{tabular}
\end{table}

\subsection{Impact of Different Driving Styles and Driving Scenario Difficulty on Decision Tree Structure}


We further investigate how driving styles and scenario difficulty influence the structure of decision trees. Figure~\ref{fig:COMPARE_TREE} compares the decision trees generated under prompt settings for conservative and aggressive driving styles, using a fixed environment configuration of 4 lanes with a vehicle density of 2.00. It can be observed that the decision tree corresponding to the conservative driving style is relatively shallow and structurally simple, whereas the one for the aggressive style exhibits greater depth and structural complexity. This is because, under the aggressive tactic, the large language model actively seeks ways to complete the driving task as efficiently and quickly as possible, which increases the likelihood of accidents. As a result, the model engages in more nuanced reasoning about the driving environment, generating a richer set of conditional nodes to handle extreme or complex driving situations.

\begin{figure}[htbp]
    \centering
    \includegraphics[width=0.8\textwidth]{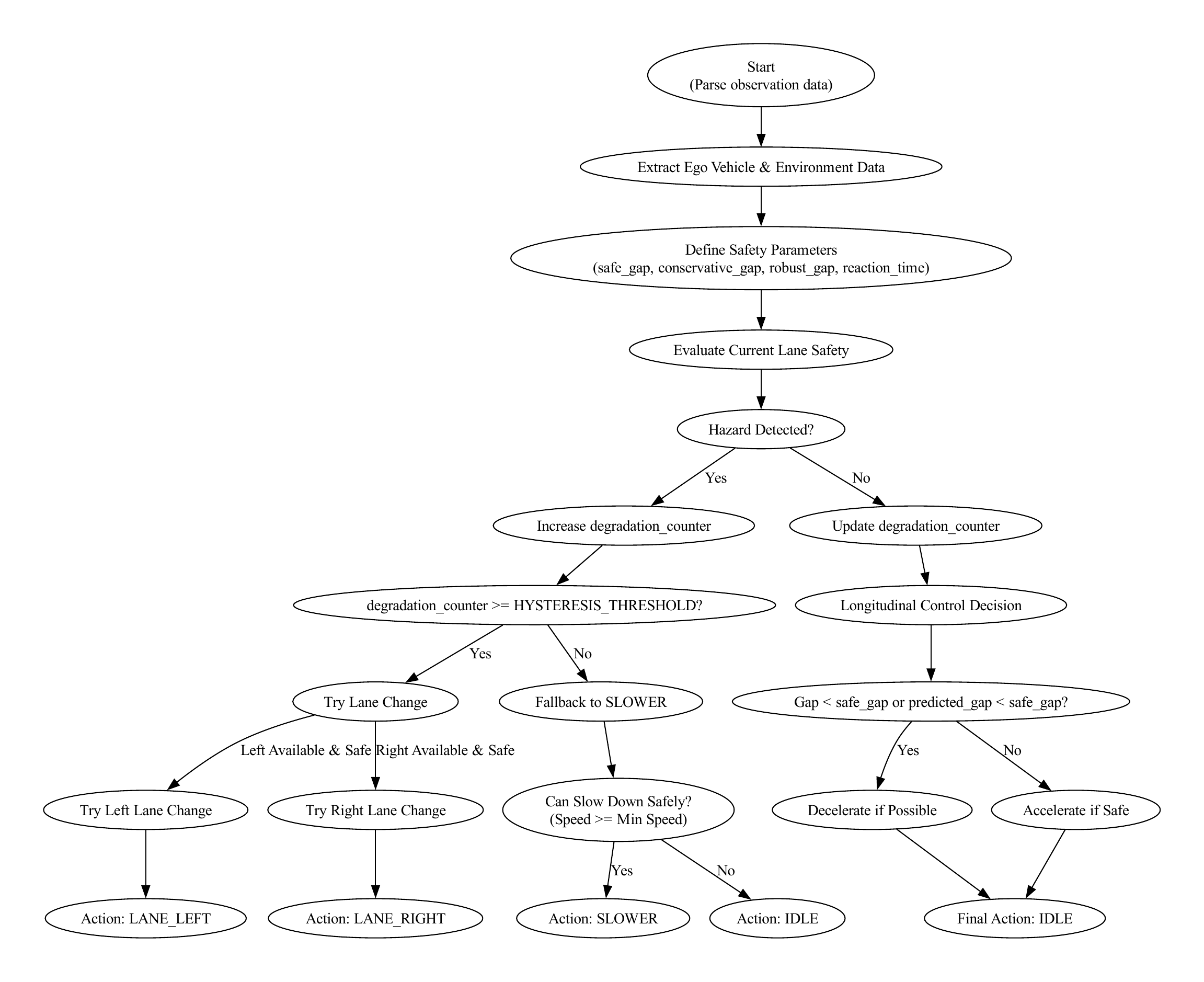}
    \caption{Decision Trees Obtained from Conservative Policies}
    \label{fig:COMPARE_TREE}
\end{figure}

\begin{figure}[htbp]
    \centering
    \includegraphics[width=0.8\textwidth]{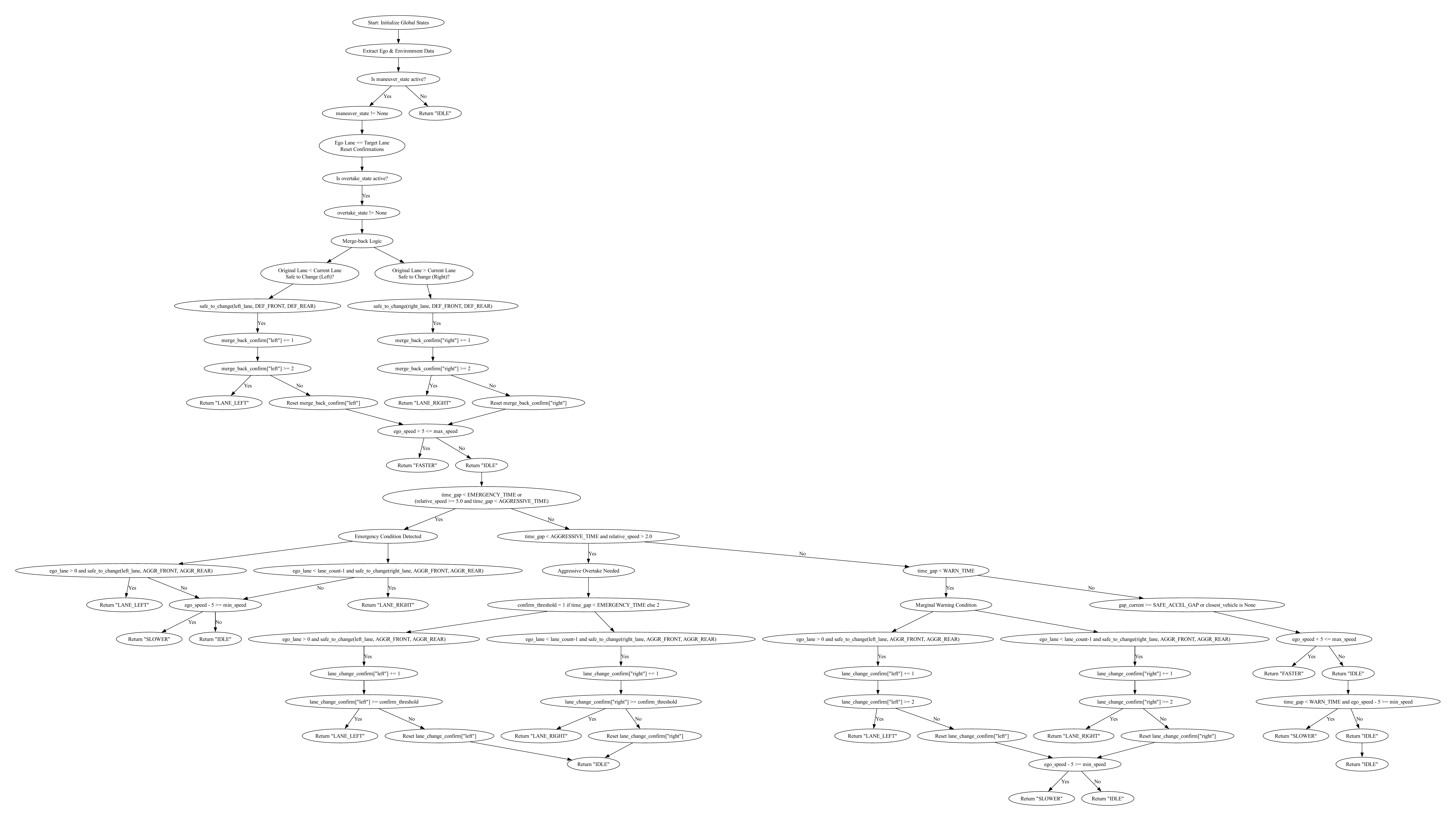}
    \caption{Decision Trees Obtained from Aggressive Policies}
    \label{fig:COMPARE_TREE}
\end{figure}


Furthermore, we analyze how varying levels of driving scenario difficulty affect the structure of the decision trees. To amplify structural differences across scenarios, we use the aggressive driving tactic and conduct experiments under configurations of 4 lanes with vehicle densities of 0.75, 1.00, and 1.25. A relatively lower vehicle density is selected to ensure that the learned policies remain largely collision-free during training. The structural parameters of the resulting decision trees are summarized in Table~\ref{tab:DIFFICULTY_TREES}. We observe that as the difficulty of the driving scenario increases, the decision trees produced by \textbf{ADRD} become progressively more complex. Notably, when the vehicle density reaches 1.25, the depth of the decision tree sharply increases to 34. This indicates that as driving conditions become more challenging, the large language model must make finer distinctions among different traffic situations in order to generate decisions that are both safe and consistent with the aggressive driving target.

\begin{table}[h]
    \centering
    \caption{Summary of decision tree characteristics under different vehicle densities.}
    \label{tab:DIFFICULTY_TREES}
    \begin{tabular}{c c c c}
        \toprule
        \textbf{Vehicle Density} & \textbf{Number of Nodes} & \textbf{Number of Branches} & \textbf{Decision Tree Depth} \\
        \midrule
        Low (0.75) & 49 & 56 & 10 \\
        Medium (1.00) & 61 & 60 & 12 \\
        High (1.25) & 65 & 68 & 34 \\
        \bottomrule
    \end{tabular}
\end{table}

\section{Conclusion}


In this paper, we explore the use of large language models to generate decision trees for autonomous driving decision-making. We model each component involved in the task of autonomous driving decision tree generation and propose the \textbf{ADRD} framework, which consists of the Information Module, the Agents Module, and the Testing Module. This framework integrates multi-source information such as driving scenarios and traffic regulations, enabling the generation of effective decision trees for autonomous driving. Moreover, \textbf{ADRD} supports iterative improvement of these decision trees through automated testing and refinement.

Extensive experimental results demonstrate that \textbf{ADRD} outperforms not only traditional reinforcement learning methods but also Dilu, a knowledge-driven approach that previously achieved state-of-the-art performance on highway-env. \textbf{ADRD} enables autonomous driving decisions that are high-performing, fast in response, and highly interpretable. Furthermore, it exhibits a general and comprehensive understanding of driving scenarios, generating flexible and easily modifiable decision trees, thereby showing great potential for application in real-world autonomous driving systems.


\bibliographystyle{unsrt}  
\bibliography{references}

\newpage
\appendix

\section{Prompt Templates}
\subsection{Planner}
\vspace{-\baselineskip}

\begin{figure}[H]
    \centering
    \includegraphics[width=0.75\textwidth]{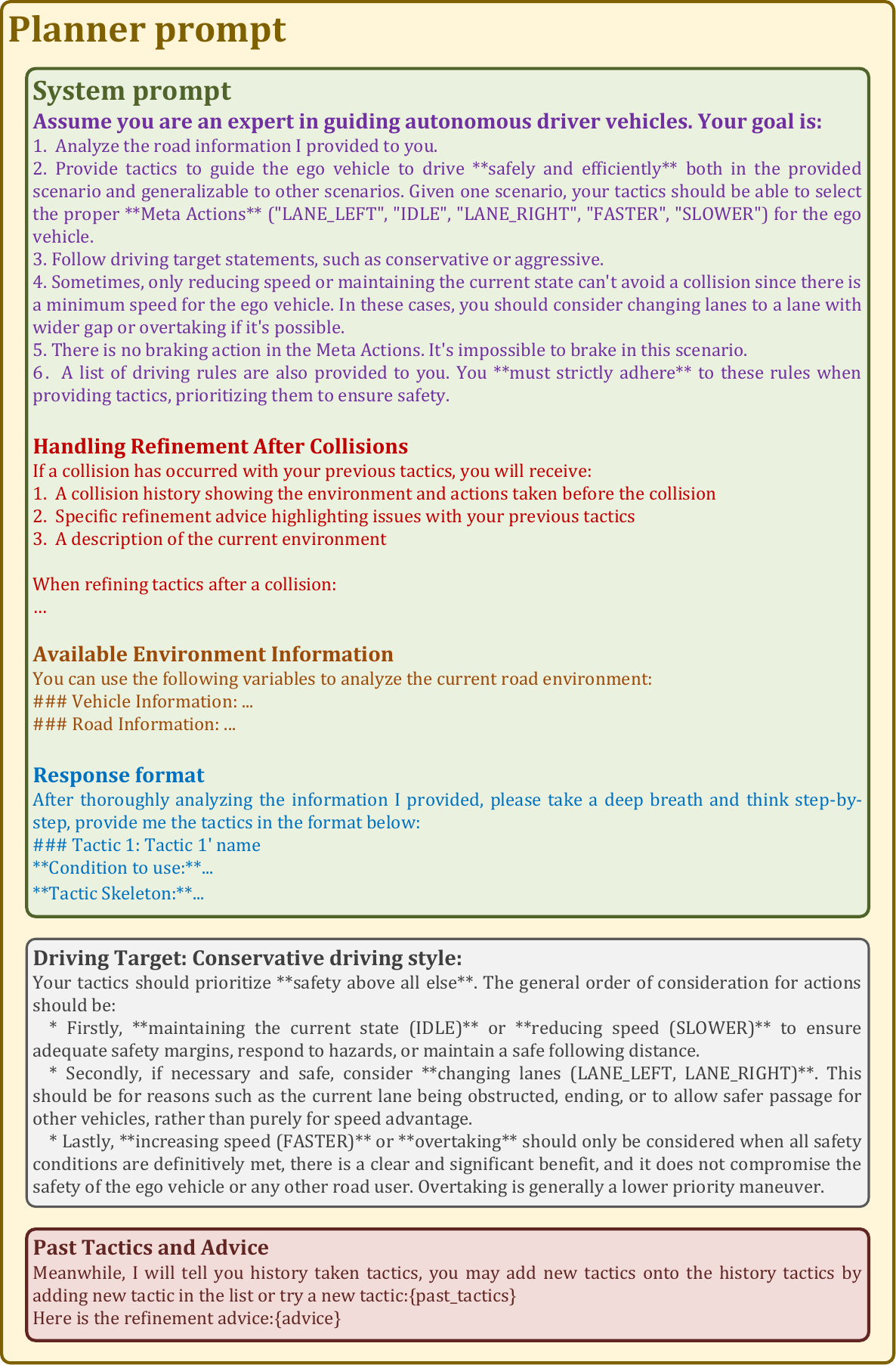}
    \caption{Example Prompt Template of \textbf{Planner}.}
    \label{fig:example}
\end{figure}

\newpage
\subsection{Coder}
\vspace{-\baselineskip}

\begin{figure}[H]
    \centering
    \includegraphics[width=0.75\textwidth]{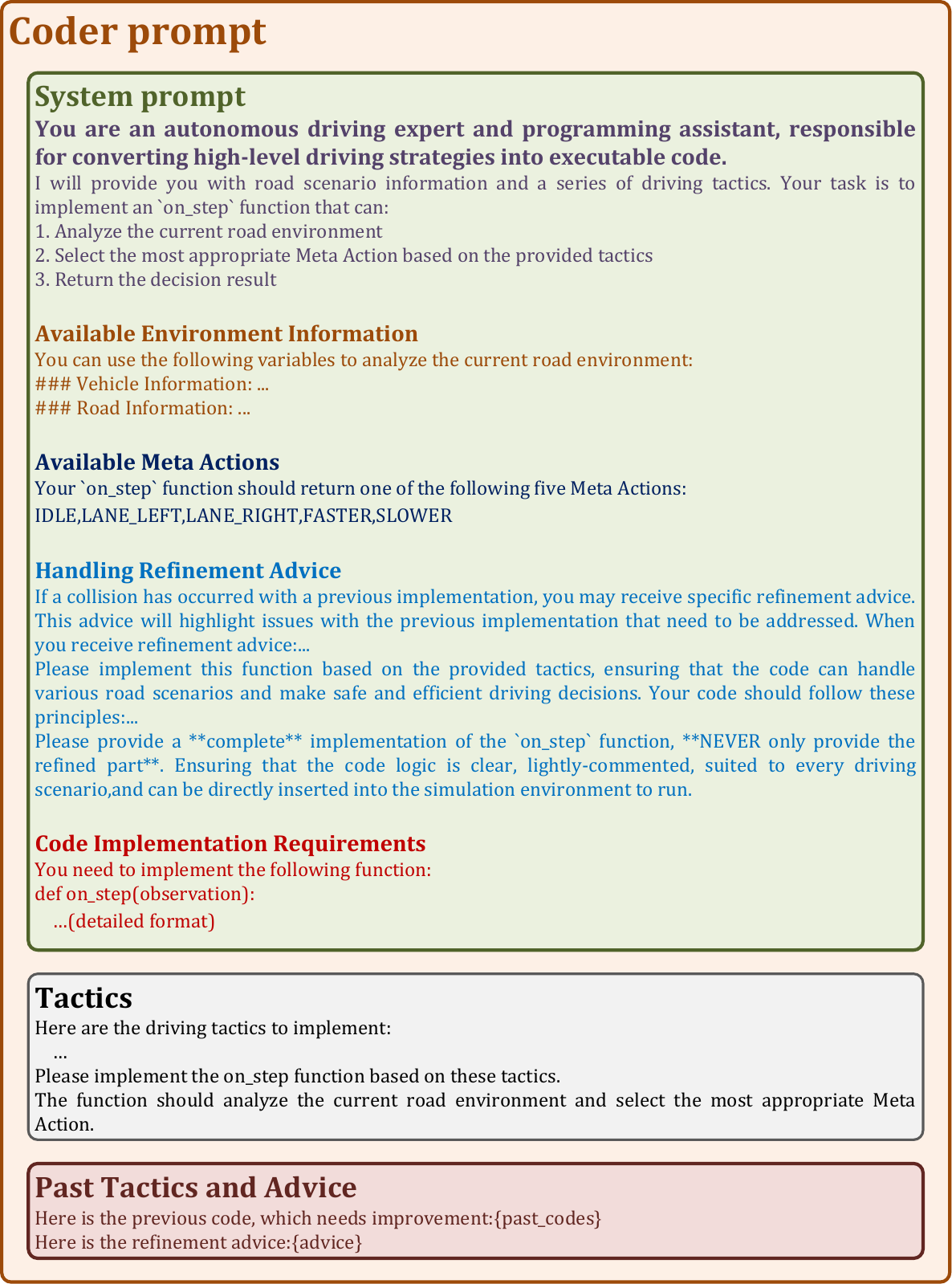}
    \caption{Example Prompt Template of \textbf{Coder}.}
    \label{fig:example}
\end{figure}

\newpage
\subsection{Summarizer}

\begin{figure}[H]
    \centering
    \includegraphics[width=0.75\textwidth]{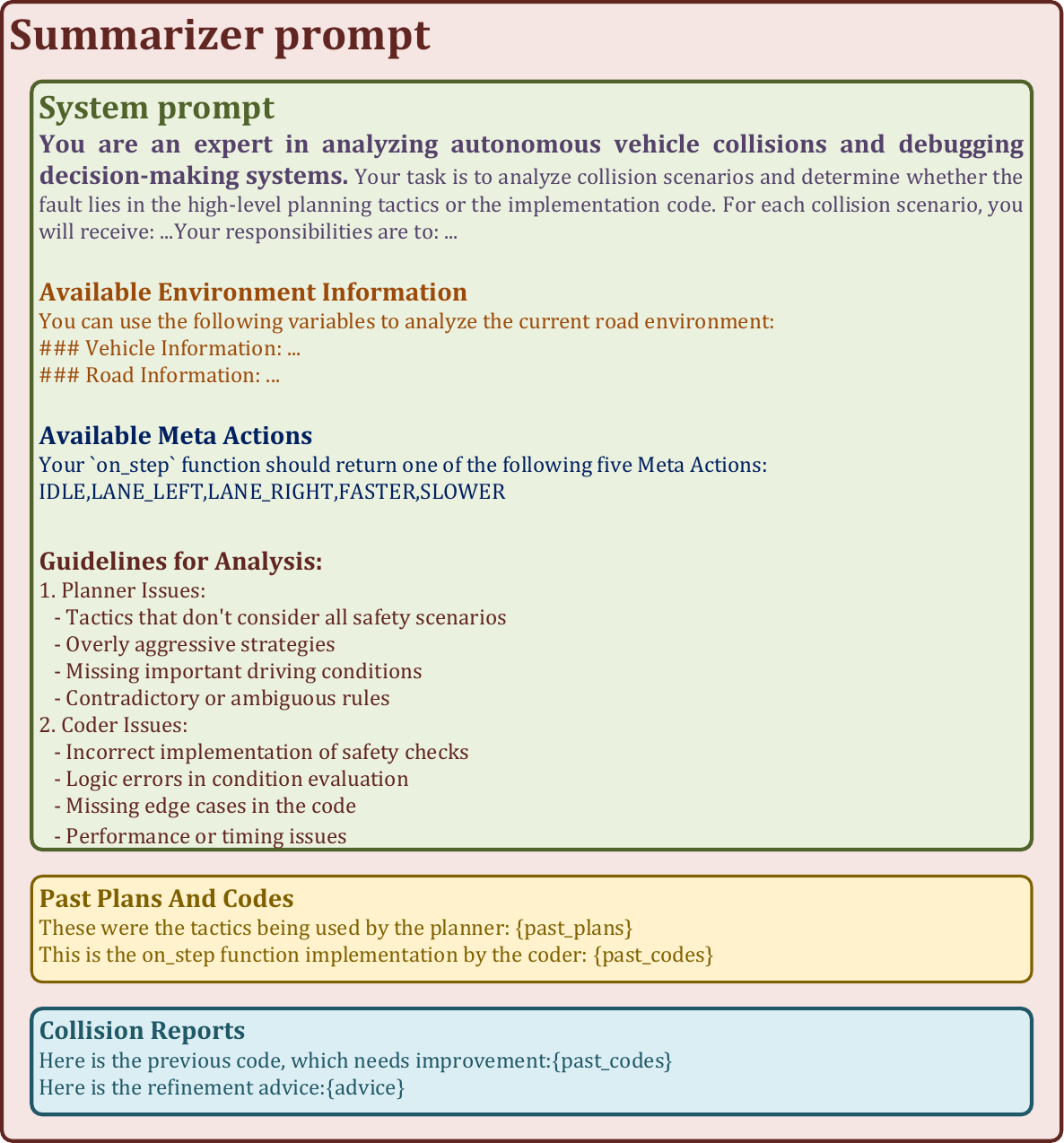}
    \caption{Example Prompt Template of \textbf{Summarizer}.}
    \label{fig:example}
\end{figure}

\end{document}